\documentclass{article} % For LaTeX2e
\usepackage[utf8]{inputenc}
\usepackage{iclr2022_conference,times}
\usepackage{graphicx}
\usepackage{xcolor}

\usepackage{amsmath,amsfonts,bm}

\def\eqref#1{equation~\ref{#1}}
\def\1{\bm{1}}

\DeclareMathAlphabet{\mathsfit}{\encodingdefault}{\sfdefault}{m}{sl}
\SetMathAlphabet{\mathsfit}{bold}{\encodingdefault}{\sfdefault}{bx}{n}

\usepackage{hyperref}
\usepackage{url}

\title{System Analysis for Responsible Design of Modern AI/ML Systems\thanks{DISTRIBUTION STATEMENT A. Approved for public release. Distribution is unlimited.
This material is based upon work supported by the Under Secretary of Defense for Research and Engineering under Air Force Contract No. FA8702-15-D-0001. Any opinions, findings, conclusions or recommendations expressed in this material are those of the author(s) and do not necessarily reflect the views of the Under Secretary of Defense for Research and Engineering. © 2021 Massachusetts Institute of Technology.
Delivered to the U.S. Government with Unlimited Rights, as defined in DFARS Part 252.227-7013 or 7014 (Feb 2014). Notwithstanding any copyright notice, U.S. Government rights in this work are defined by DFARS 252.227-7013 or DFARS 252.227-7014 as detailed above. Use of this work other than as specifically authorized by the U.S. Government may violate any copyrights that exist in this work.}}

\author{Virginia H. Goodwin, Rajmonda S. Caceres \\
MIT Lincoln Laboratory\\
\texttt{\{virginia.goodwin,rajmonda.caceres\}@ll.mit.edu} \\}

\iclrfinalcopy % Uncomment for camera-ready version, but NOT for submission.
\begin{document}

\maketitle
\begin{abstract}
The irresponsible use of ML algorithms in practical settings has received a lot of deserved attention in the recent years. We posit that the traditional system analysis perspective is needed when designing and implementing ML algorithms and systems. Such perspective can provide a formal way for evaluating and enabling responsible ML practices. In this paper, we review components of the System Analysis methodology and highlight how they connect and enable responsible practices of ML design.
\end{abstract}

% Sections
\section{Introduction}\label{sec:Intro}
In recent years the Machine Learning (ML) community has begun to address serious issues with lack of responsible use of ML algorithms~\cite{lazovich2022measuring, molnar22,DBLP:journals/corr/abs-1904-03257}. 
%We assert that the practice of Systems Analysis is well-suited to guiding researchers and developers on the path of doing appropriate impact assessments.
%Machine Learning 
Researchers are starting to address the utility and possible overuse or misuse of standard ML challenge data sets, such as PASCAL VOC or ImageNet in the computer vision (CV) domain, and GLUE/SuperGLUE in Natural Language Processing (NLP)~\cite{raji2021ai}. Too frequently, these common and restricted benchmark data sets are taken as the end-all-be-all of algorithm performance measurement, and yet our experience deploying ML algorithms in fielded systems has shown time and time again that this method leaves a lot to be desired in the context of real-world performance on novel data.
We observe that even the algorithms that achieve best in class performance against the benchmark data do not necessarily result in sufficient performance in novel environments. As Green says,~\cite{Green20194} ``... although most people talk about machine learning’s ability to predict the future, what it really does is predict the past." Formalism and mitigation approaches have been offered for specific problems and algorithms (e.g., classification)~\cite{AIrobustness,DBLP:journals/corr/KatzBDJK17,kleinberg2016,lundberg2017,suresh2019,suresh2021}. However, these are currently only applicable in very limited circumstances. 

We believe that taking a traditional Systems Analysis approach to ML assessment offers a coherent pathway out of this problem, which will result in better outcomes for both algorithm development and performance in real-world systems. Current ML system development too often overlooks a critical question addressed by Systems Analysis, i.e., what are the real-world costs of false alarms and missed detections? This includes costs to the operating entity, such as the cost of missing an important detection and the cost of responding to false alarms, but as we have already seen too many times, ethical developers and users of ML systems must also consider the costs to the subjects of the ML system. Already we have seen systems deployed for operational use with no accounting for how false alarms in particular will impact people's lives. When the ML system is making decisions about who deserves an interview~\cite{Dastin18}, where police should spend more time patrolling~\cite{Green20194}, identifying and even predicting criminal behavior~\cite{Hao19}, and making sentencing recommendations~\cite{Heaven20}, the cost to the individuals impacted by false alarms is profound. ``Move fast and break things" is a terrible and immoral operating philosophy when the ``things" being broken are people's lives.

In this paper, we first define Systems Analysis; %which admittedly can be somewhat squishy; 
we then go on to draw parallels between the main points of Systems Analysis and how they can apply specifically to ML systems.

%There is one particular paper that talks about the difference between bias in classification and bias in decision systems that we should reference.
%System level metrics of evaluation (or lack thereof): MLSys: The New Frontier of Machine Learning Systems: https://arxiv.org/abs/1904.03257

\section{What is System Analysis?}\label{sec:SAdef}

Historically, the practice of Systems Analysis is closely linked with that of Systems Engineering. Systems Engineering began at Bell Labs in the early 1900s and became a significant field in its own right during the industrialization driven by WWII \cite{Buede16}. RAND Corporation is also credited with the initiation of Systems Analysis as a separate field in the fifties \cite{Buede16}, although systems analysis was being done as part of systems engineering earlier.

Systems analysis is often a squishy topic, as it is hard to generalize across a wide array of types of systems, including hardware, software, human-in-the-loop (HITL) processes, geospatial and temporal processes. For the purposes of this discussion we break Systems Analysis down into four main steps:

\textbf{1. Scoping the Problem Space.} This first step includes understanding both the technical challenges and also the full environment and constraints of who is going to be using the final system. This should include understanding the concept of operations (CONOPS), which is, how the operator will use the system, what rules circumscribe their choices and actions, what response resources they have, and the cost of mistakes. This is ideally where metrics of performance and success should be defined. The most common metrics for the system are the false alarm and missed detection rates that are considered tolerable by the operator. Scoping and properly defining the problem space is typically the bulk of systems analysis work.

\textbf{2. Scoping the Solution Space.} This second step involves understanding or specifying out what possible solutions exist, frequently termed ``the art of the possible''~\cite{Delaney2015}. This may include looking at different hardware and sensor options, as well as understanding what constraints are placed on your solutions by the operator's available responses. This may also include looking at HITL trade-offs, such as whether a user will be available to provide input or if the system must perform independently.

\textbf{3. Creating Candidate Solutions.} Once we understand the range of possible solutions to a problem, the next step is to build one or more solutions. There are inevitably challenges that must be overcome in making something actually work that aren't addressed in the higher-level scoping of the solutions space. Furthermore, once a prototype exists, the human element (CONOPS, response, etc.) inevitably changes when the theorized CONOPS don't work as well as expected, or new ways to use the system are discovered.

\textbf{4. Assessing Performance.} Here we circle back to the metrics of performance specified in Step 1, and assess how well the candidate solution performs relative to those metrics. These should include not just technical performance, but how well the system improves operator performance, what are the cost trade-offs of mistakes (both false alarms and missed detections) and also what unintentional effects does the system have.
    
As ML algorithm development is currently practiced, many algorithms are published, and subsequently reused in a multitude of different contexts, where the basis of algorithm training, and/or the metrics of performance, are all built on a published benchmark data set. In these instances, developers are a priori limiting themselves in both Items 1 and 4 of the above list. They are not adequately scoping the problem space, because that space may not be properly captured in the pre-collected and cleaned benchmark data set, and if they simply rely on the traditional benchmark performance metrics, they are not looking at whole-system performance. We believe that if the above rubric is used from the beginning of algorithm development, this will result in better outcomes for system performance in final systems.

\section{Parallels to AI/ML Systems}\label{sec:MLparallels}
In this section we take a closer look at each step in Systems Analysis, and make connections to the machine learning process.

\subsection{Scoping the Problem Space}\label{subsec:Step1}
As noted above, the vast majority of Systems Analysis is performed in Step 1. We can unpack this into the following pieces: 1)
Input data and associated technical challenges, 2) Output results and associated costs, 3) Constraints on the system.

\textbf{Input Data and Associated Technical Challenges.} Whereas most ML algorithms rely on published data sets to train and against which to demonstrate performance, deployed systems are ingesting data ``in the wild'', where there is typically lots of noise as well as larger data distribution issues, such as uneven class distribution, or distributions may shift over time, as the environment changes. A deep understanding to the technical underpinnings of the data to be analyzed is critical to identify possible solutions, and possible sources of error. In addition to the details of the collected data, there are typically technical challenges related to processing power, input data resolution or quality, and myriad data processing steps that need to be addressed, such as the raw data from a sensor may be very far from what is expected as input to the ML system. Each of these steps may introduce some bias or error in the process that needs to be understood.
        
\textbf{Output Results and Associated Costs.} Here we need to understand the system-level output that is required, whether it is an alert to an operator, a classification result, an image, etc., and most importantly, how those results are acted on, and what are the costs associated with errors to the system-level outputs. The cost of false alarms and missed detections is possibly the most critical piece of the full Systems Analysis. In most systems with an HITL component, operators like to think about False Alarm Rates and Missed Detection Rates, i.e. how many false alarms and missed detections can the operator tolerate per hour, per day, etc. Critically, in addition to the costs to the operator of responding to false alarms and not responding to missed detections, there is the cost to the subject of the false alarm or missed detection. For example, if the response to a false alarm is a law enforcement intervention, there is a critical cost to the person who is being erroneously targeted by the response team. For emerging ML-based systems, these costs are only beginning to be understood and considered.
        
\textbf{Constraints on the System.} Closely related to the costs of false alarms and missed detections, there are typically constraints on the system, both technical and operational. The ML community is familiar with technical constraints such as processing power, data quality, data transmission rates, etc. On top of that, for an operational system, we must also consider operational constraints such as human resources, laws, rules, and social norms, and other ``external'' system constraints. In practice, many technically functional systems don't get used because they break the operational constraints of the whole system.

It is critical to clearly define the full scope of the problem space before we embark on proposing a solution. Details that will emerge in scoping the solution space, such as choosing a specific algorithm or learning objective, creating and assessing performance metrics, and data preprocessing and handling, will all have an impact on the performance of the final system.

\subsection{Scoping the Solution Space}\label{subsec:Step2}
Scoping the solution space is essentially doing a survey of existing and potential future technologies, and understanding how various components might be put together to build out a system that will address the user needs. In the context of ML, given the planned input data per Section \ref{subsec:Step1}, there may be various different ML algorithms that would all result in the desired output, or potentially multiple algorithms may be used in ensemble to get to the final desired result. Frequently, the type or quality of available input data, and other system-level constraints, will restrict the space of potential solutions. The challenge here in the context of the whole system is understanding how error or bias may be introduced first from the input data, then potentially propagated and magnified through the processing of the ML algorithms. 

Additionally, as part of scoping out potential solutions, there may be various different options for exactly how the output results are generated or presented. For example, classification system may output different classes with associated confidence scores, and then there are multiple options at the system level to assist an operator in making a decision. The system can either present the top N results if there is an operator available to choose among the results, or the system can assume less human resources and output the top single classification as the definitive decision. These types of performance trade-offs are a large part of the the cost-benefit balancing necessary for developing useful systems.

\subsection{Create One or More Candidate Solutions}\label{subsec:Step3}
Based on the outcome of Steps 1 and 2, one or more potential solutions may be proposed. For example, we may look at a more computationally-intensive solution that gives better performance, vs. a lighter-weight solution that gives slightly worse performance but at lower computational cost, and we may seek to mitigate performance shortfalls at another level of the system, such as by presenting the operator with multiple choices vs. one choice.

The challenge in this stage is to rigorously understand and assess the performance of each system component, while never losing sight of the goals of the full system.

\subsection{Assessing System Performance}\label{subsec:Step4}
Finally, here we circle back to Step 1, where we discussed the overall system performance metrics and costs. It is critical to assess the full system performance that an operator will experience, and then trace the performance back to individual component performance. This whole process is usually iterative, where we get a first base-line performance result and then go into the system, perhaps even re-architecting to include more or different algorithms to achieve the desired overall system performance. Here we can also make trade-offs of requiring more HITL time (if available) to offset algorithm performance.

\subsection{Systems Analysis Without a System}
While all of the preceding has rested on the presumption that a system exists, that operators of the system exist, and that developers can understand all relevant aspects of the above, it is also true that ML systems may be developed in the absence of an existing system from which to glean these critical details. In the absence of knowing a priori how an algorithm will be deployed, the alternative solution is documentation. It is imperative that the developers rigorously document the training data and trained model that they create, including both the intended use cases and inappropriate or out of scope applications. As noted in~\cite{diuRAI}, the use of rigorous documentation such as Data Sheets~\cite{gebru2020datasheets} and Model Cards~\cite{MitchellModelCards2019} is not only best practice for documentation, but also benefits the final system by eliciting critical questions and issues early in the design process. In addition, it is critical to create a measurement framework~\cite{holstein2018} by identifying points within the system at which results can be assessed at different levels (data and sensing, algorithmic design, human-AI interaction, and system/application level), and create `hooks' in the software to enable continuous or periodic reassessment of performance over time.

\section{Conclusion}\label{sec:conclusion}
While the additional work required to do systems analysis is non-negligible, we believe that the benefits of doing this level of analysis at each stage of an AI/ML system development are quite significant to the final performance of the system. Many people in the ML community are deeply concerned about the misuse of ML systems in real-world settings with real costs to individuals and society. Engaging in systems analysis from the very start of the effort can call out and mitigate against these harms, in particular by making explicit correct and incorrect use of available data, as well as external constraints and costs associated with such a system. There is also benefit in clearly documenting decision-making responsibility and CONOPS between the ML algorithm and human users of the system to clarify accountability and ownership. In many cases we recognize that ML algorithm development is divorced from a future real-world system, and therefore full system metrics cannot be evaluated. In these cases, we believe that rigorous documentation of the data and model training can bridge the gap, such that future developers are able to make informed decisions regarding how appropriate it is to apply the data or model to their specific problem. 

\bibliographystyle{iclr2022_conference}
\bibliography{references}

\end{document}